\documentclass[letterpaper]{article}
\usepackage{textcomp}
\usepackage{amsmath}
\usepackage[submission]{aaai2027}  
\makeatletter
\gdef\showauthors@on{T}  
\gdef\copyright@on{}     
\makeatother
\usepackage[hyphens]{url}  
\usepackage{xcolor}
\definecolor{citeblue}{RGB}{0,82,204}
\usepackage{graphicx} 
\usepackage{natbib}  
\usepackage{caption} 
\usepackage{algorithm}
\usepackage{algorithmic}
\usepackage{newfloat}
\usepackage{multirow} 
\usepackage{amsmath}
\usepackage{amsfonts}
\usepackage{listings}
\usepackage{pifont}
\usepackage{array}
\usepackage{xcolor}
\definecolor{citeblue}{RGB}{0,82,204}
\DeclareCaptionStyle{ruled}{labelfont=normalfont,labelsep=colon,strut=off} 
\floatstyle{ruled}
\newfloat{listing}{tb}{lst}{}
\floatname{listing}{Listing}
\usepackage{booktabs}
\usepackage{amssymb}
\title{WaveFreqAnchor: Wave-Structural Anchoring and Frequency Correction Diffusion for Training-Free Face Restoration}
\author {
    Zelin Du\textsuperscript{\rm 1}\textsuperscript{$\dagger$},
    Wenjie Li\textsuperscript{\rm 2}\textsuperscript{$\dagger$},
    Zhengxue Wang\textsuperscript{\rm 3},
    Juncheng Li\textsuperscript{\rm 4},
    Cailing Wang\textsuperscript{\rm 1},
    Guangwei Gao\textsuperscript{\rm 3}\thanks{Corresponding author,   $\dagger$Equal contributions.}
}

\affiliations {
    \textsuperscript{\rm 1} Nanjing University of Posts and Telecommunications  
    \textsuperscript{\rm 2} Beijing University of Posts and Telecommunications\\
    \textsuperscript{\rm 3} Nanjing University of Science and Technology
    \textsuperscript{\rm 4} East China Normal University \\
    \{klive0417, lewj2408\}@gmail.com,
    \{zxwang, gwgao\}@njust.edu.cn,
    jcli@cs.ecnu.edu.cn,
    wangcl@njupt.edu.cn
}

\begin{document}
\maketitle
\begin{abstract}

Diffusion-based face restoration that adjusts the sampling trajectory of pre-trained diffusion models has achieved remarkable progress. However, existing approaches provide insufficient constraints during reverse diffusion, causing identity-related structural drift and degraded fidelity under severe degradations. To address this, we propose WaveFreqAnchor, a training-free framework based on Wave-Structural Anchoring and Frequency Correction Diffusion. Specifically, Anchor-Space Wave-Structural Guidance (ASWG) constrains facial structures through anisotropic wave-response consistency, while Multi-scale Wavelet-Fourier Injection (MWFI) aligns the predicted low-frequency subband with the observation by replacing its phase, correcting inconsistencies accumulated during reverse diffusion. For real-world scenes, we further introduce Subband High-Frequency Enhancement (SHE), which performs bounded, spatially masked refinement on the predicted high-frequency subbands to recover fine facial details under unknown compound degradations. Together, these designs effectively preserve facial identity while restoring sharp and realistic facial details. Extensive experiments show that our method consistently outperforms existing methods, achieving high-quality and high-fidelity face restoration.

\end{abstract}

\section{Introduction}

\begin{figure*}[t]
\centering
\includegraphics[width=\textwidth]{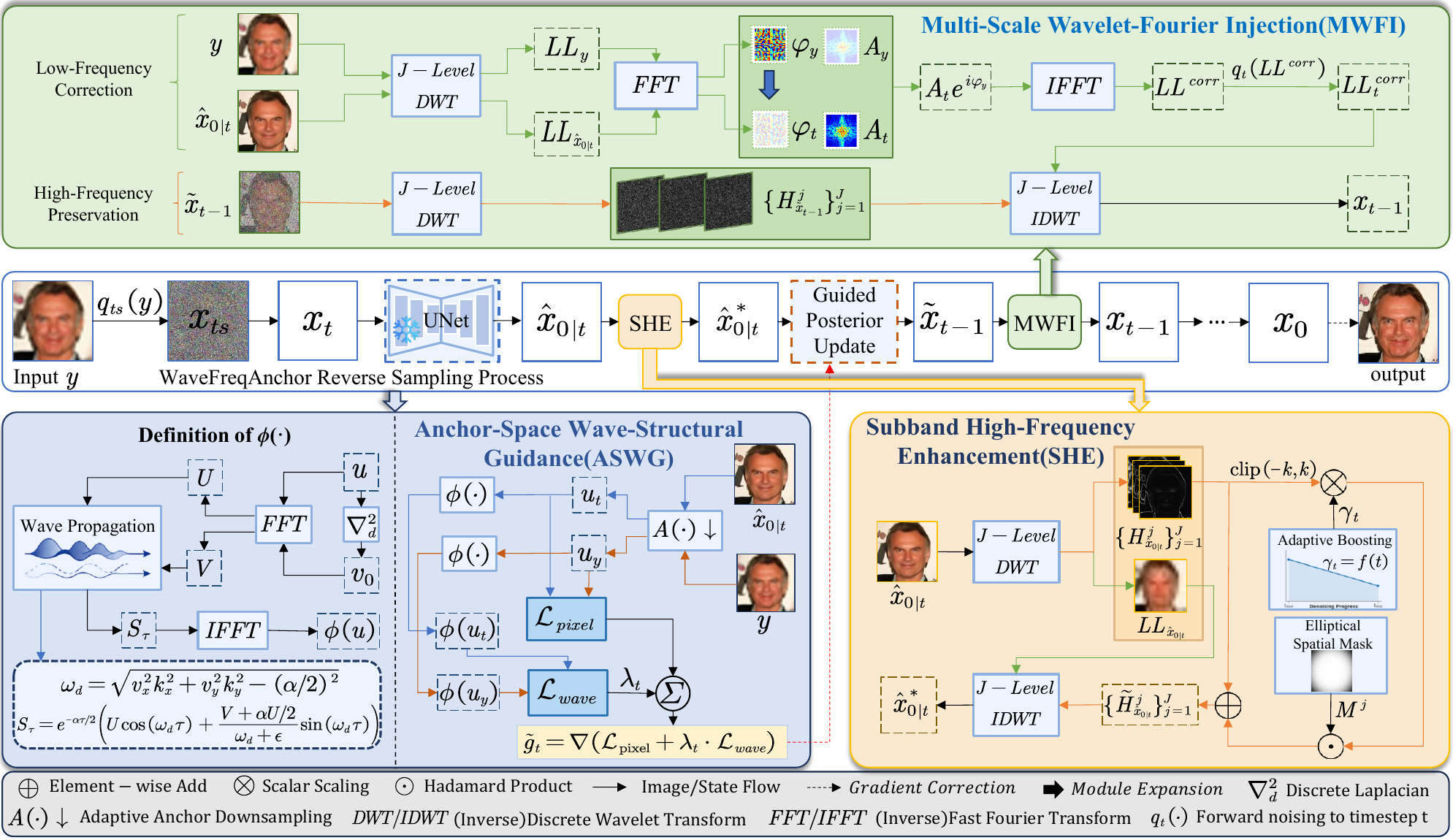}
\caption{Overview of our WaveFreqAnchor. Given the input observation, ASWG provides gradient guidance for posterior sampling, while MWFI corrects the predicted low-frequency wavelet subband using the observed phase and preserves the high-frequency subbands from the guided update. For real-world inputs with missing high-frequency details, SHE further refines the predicted high-frequency subbands before the posterior update.}
\label{fig:framework}
\end{figure*}

Face restoration~\cite{li2025survey} aims to recover a high-quality face image from a degraded input. Under severe degradations, only coarse facial structures and limited identity cues remain, making the restoration problem highly ill-posed and often leading to plausible yet identity-inconsistent results. Existing methods leverage generative facial priors~\cite{wang2021towards}, discrete codebooks~\cite{zhou2022towards}, and diffusion priors~\cite{lin2024diffbir, wang2025osdface} to constrain the solution space. However, even when producing visually realistic results, they often fail to faithfully preserve the observed facial structures under severe degradations. Achieving both realistic detail synthesis and structural fidelity therefore remains a fundamental challenge.



Pre-trained diffusion models~\cite{yue2024difface} can generate realistic facial details for severely degraded inputs, providing an effective balance between perceptual quality and restoration fidelity. Diffusion-prior-based methods, such as DiffBIR~\cite{lin2024diffbir}, OSDFace~\cite{wang2025osdface}, and FiDeSR~\cite{kim2026fidesr}, rely on task-specific training, incurring substantial computational overhead. In contrast, posterior sampling methods~\cite{chung2023diffusion,zheng2026image} avoid retraining by incorporating degradation operators during reverse sampling. However, under severe degradations, the observed inputs often contain insufficient structural cues. This challenge is further exacerbated in real-world scenarios, where unknown degradation operators further prevent accurate degradation modeling and cause the reverse sampling trajectory to drift from the underlying facial structure. Therefore, for faithful face restoration, the key to posterior sampling is to effectively exploit limited observations to constrain reverse sampling.


Wavelet-domain representations~\cite{mallat1989theory} provide effective structural cues for face restoration by decomposing images into low- and high-frequency subbands~\cite{li2024efficient}. The low-frequency subband preserves the coarse facial structure, while its Fourier phase~\cite{oppenheim1981importance} encodes critical spatial information that remains relatively stable even under severe degradations. In contrast, the high-frequency subbands capture directional edges and fine facial textures, but their observed responses are often weak or unreliable. Consequently, insufficient low-frequency constraints can cause the restored face to deviate from the observed structure, whereas directly propagating or uniformly enhancing the observed high-frequency responses may amplify unreliable details. Therefore, low-frequency structures should remain aligned with the observations, while high-frequency details are better recovered through guided diffusion updates with additional controlled refinement of high-frequency information for severe degradations.


Based on the above observations, we propose WaveFreqAnchor, a training-free diffusion framework for face restoration. Starting from a noised observation, it guides a frozen diffusion model through wave-structural anchoring and frequency correction. Anchor-Space Wave-Structural Guidance (ASWG) projects the predicted clean image to a low-resolution anchor space and enforces observation-consistent facial structures using pixel consistency and Laplacian-initialized anisotropic wave-response constraints. Multi-scale Wavelet-Fourier Injection (MWFI) preserves the predicted low-frequency amplitude, aligns its phase with the observation, and retains the guided high-frequency subbands. For synthetic super-resolution, the known degradation model provides an explicit low-resolution anchor, allowing the guided diffusion process to recover missing high-frequency details without additional refinement. For real-world restoration, where unknown degradations often leave facial details under-recovered, a frequency-ratio router adaptively selects the sampling configuration, and Subband High-Frequency Enhancement (SHE) performs bounded, spatially masked refinement on the predicted high-frequency subbands, while low-frequency correction is handled by MWFI. In summary, our contributions can be summarized as follows:
\begin{itemize} 
\item We proposed that ASWG imposes constraints on key facial structures and that MWFI corrects the low-frequency subbands in the facial predictions, thereby effectively constraining facial identity features.
\item For real-world degradations, we introduce a frequency-ratio routing scheme with input-dependent sampling configurations, and a SHE with bounded spatial masking to optimize the recovery of lost high-frequency details.
\item Experiments show that our method effectively improves the identity similarity of face restoration without training, particularly under severe degradation, while delivering excellent visual perceptual quality.
\end{itemize}

\section{Related Work}

\noindent\textbf{Generative Priors for Face Restoration.}
Face restoration under severe degradation often relies on generative priors. PULSE~\cite{menon2020pulse} searches the latent space of a pretrained GAN. Feed-forward methods incorporate generative facial priors~\cite{yang2021gpen,wang2021towards}, vector-quantized codebooks~\cite{gu2022vqfr,zhou2022towards}, key-value priors~\cite{wang2022restoreformer}, or prior-based latent transformations~\cite{xie2024pltrans}. Recent diffusion methods use a diffusion-based degradation remover followed by an enhancement module~\cite{wang2023dr2}, combine a restoration module with diffusion-based generative refinement~\cite{lin2024diffbir}, bridge a restoration backbone to a pretrained diffusion prior through a transition distribution~\cite{yue2024difface}, perform one-step restoration~\cite{wang2025osdface}, or adapt the starting timestep and spatial guidance to estimated blur levels~\cite{do2025dynfacerestore}. These methods rely on image-specific optimization or task-specific training. Our method instead guides a frozen face diffusion prior during inference without task-specific training.

\noindent\textbf{Diffusion Priors for Restoration.}
Pretrained diffusion priors have also been used for inverse restoration. DDRM~\cite{kawar2022denoising} and DDNM~\cite{wang2023zeroshot} enforce consistency under known linear operators. DPS~\cite{chung2023diffusion} guides reverse sampling with likelihood gradients. RED-Diff~\cite{mardani2024variational} formulates posterior inference through variational optimization, whereas P2L~\cite{chung2024prompt} jointly optimizes prompt, latent, and pixel variables through alternating minimization. DAPS~\cite{zhang2025improving} uses decoupled noise annealing, and SubDAPS++~\cite{zheng2026image} extends it with dynamic-resolution diffusion priors. These methods generally assume a specified degradation operator and do not explicitly constrain observation-supported facial structure. PGDiff~\cite{yang2023pgdiff} uses predefined image properties to guide pretrained diffusion priors, whereas SSDiff~\cite{li2025self} constructs staged guidance from pseudo-references. PASDiff~\cite{ni2026pasdiff} combines physics-aware photometric constraints with structural injection from an off-the-shelf face prior for low-light face restoration. In contrast, our method couples anchor-space wave-structural guidance with frequency-specific correction to keep reverse sampling tied to the facial structure supported by the observation.

\noindent\textbf{Frequency-Domain Restoration.}
Frequency-domain cues have been used in restoration objectives and model design. Focal Frequency Loss~\cite{jiang2021focal} emphasizes difficult spectral components during reconstruction, whereas Fourier space losses~\cite{fuoli2021fourier} directly supervise frequency-domain discrepancies. WFEN~\cite{li2024efficient} mitigates downsampling-induced feature distortion by decomposing facial features into low-frequency and high-frequency components. WaveFace~\cite{miao2024waveface} applies diffusion to the low-frequency component and uses a separate network to recover high-frequency components. Recent diffusion methods use low- and high-frequency adaptive enhancers~\cite{kim2026fidesr}, wavelet-based high-frequency guidance~\cite{yang2026hdwsr}, frequency-aligned self-distillation~\cite{choi2026framer}, or high-pass-filter-based diffusion guidance~\cite{li2026seeing}. Given that Fourier phase carries spatial structure and high-frequency subbands encode local facial details, our method leverages these cues during reverse sampling by correcting the predicted low-frequency wavelet subband with the observation phase, preserving the high-frequency subbands from the guided update, and further refining them for real-world inputs, thereby maintaining observation-consistent structure and improving local detail recovery.

\section{Methods}

\subsection{Problem Formulation and Diffusion Prior}

Let \(y\) denote the aligned and resized degraded face observation. Our method restores \(y\) with a frozen diffusion prior~\cite{ho2020denoising}, without task-specific training. For synthetic SR, we generate \(y^{lr}=\mathcal{D}_s(x^{gt})\), where \(\mathcal{D}_s\) denotes bicubic downsampling with anti-aliasing by scale factor \(s\), and upsample \(y^{lr}\) with bicubic interpolation to obtain the observation \(y\). For real-world restoration, \(y\) is the preprocessed input and its degradation operator is unknown.

Given a starting timestep \(t_s\), we initialize the reverse process from a forward-noised observation rather than from pure Gaussian noise~\cite{meng2022sdedit}:
\begin{equation}
    x_{t_s}
    =
    \sqrt{\bar{\alpha}_{t_s}}y
    +
    \sqrt{1-\bar{\alpha}_{t_s}}\varepsilon,
    \quad
    \varepsilon\sim\mathcal{N}(0,\mathbf{I}),
\end{equation}
where \(\bar{\alpha}_t\) is the cumulative noise coefficient. This initialization preserves coarse facial structure, pose, and layout while allowing the diffusion prior to recover missing details.

Following diffusion posterior sampling~\cite{chung2023diffusion}, we add a gradient correction to the reverse update. At timestep \(t\), the model predicts \(\hat{x}_{0|t}\). For real-world restoration, the high-frequency-refined estimate \(\hat{x}_{0|t}^{*}\) is used to compute the posterior mean, while \(\hat{x}_{0|t}^{*}=\hat{x}_{0|t}\) when the refinement is inactive. The guided update is
\begin{equation}
    \tilde{x}_{t-1}
    =
    \mu_t\!\left(x_t,\hat{x}_{0|t}^{*}\right)
    -
    \zeta_t \tilde{g}_t
    +
    \sigma_t\varepsilon,
    \quad
    \varepsilon\sim\mathcal{N}(0,\mathbf{I}),
\end{equation}
where \(\tilde{g}_t\) is defined as
\begin{equation}
    \tilde{g}_t=
    \frac{\nabla_{x_t}\mathcal{L}_{total}}
    {\max(\|\nabla_{x_t}\mathcal{L}_{total}\|_2,1)+\epsilon_g}.
\end{equation}
Here, \(\mathcal{L}_{total}\) is the guidance objective used by our method. It combines a pixel anchor term with an anisotropic wave-structural term, as detailed below.

\subsection{Sampling Framework}

Fig.~\ref{fig:framework} shows how the proposed corrections are incorporated into reverse sampling. ASWG computes the posterior gradient correction from the current clean estimate \(\hat{x}_{0|t}\). For real-world restoration, SHE additionally produces a refined estimate \(\hat{x}_{0|t}^{*}\). The refined estimate and the ASWG gradient are used in the guided posterior update, after which MWFI corrects the low-frequency wavelet subband while retaining the high-frequency subbands from \(\tilde{x}_{t-1}\).

The same sampling formulation is used for synthetic and real-world inputs, with different anchor targets and sampling configurations. The known synthetic scale determines the anchor resolution, wavelet level, and associated parameters. For real-world inputs, a frequency-ratio router selects between two predefined configurations. After a light Gaussian prefilter, we compute the mean absolute magnitudes of the high-frequency wavelet coefficients at the first and third decomposition levels, denoted by \(E_1\) and \(E_3\), and define
\[
    r_y = \frac{E_1}{E_3+\epsilon}.
\]
Inputs with \(r_y<\tau_r\) use the stronger configuration, and the remaining inputs use the milder configuration. The router contains no trainable component.

\subsection{Anchor-Space Wave-Structural Guidance}

\begin{table*}[t]
\centering
{\small
\setlength{\tabcolsep}{2.5pt}
\begin{tabular}{l|ccccc|ccccc|ccccc}
\toprule
\multirow{2}{*}{Method}
& \multicolumn{5}{c|}{4$\times$}
& \multicolumn{5}{c|}{8$\times$}
& \multicolumn{5}{c}{16$\times$} \\
& PSNR$\uparrow$ & SSIM$\uparrow$ & LPIPS$\downarrow$ & ID$\uparrow$ & LMD$\downarrow$
& PSNR$\uparrow$ & SSIM$\uparrow$ & LPIPS$\downarrow$ & ID$\uparrow$ & LMD$\downarrow$
& PSNR$\uparrow$ & SSIM$\uparrow$ & LPIPS$\downarrow$ & ID$\uparrow$ & LMD$\downarrow$ \\
\midrule
CodeFormer
& 27.01 & 0.789 & 0.2335 & 0.902 & 2.23
& 24.40 & 0.683 & 0.2940 & 0.739 & 2.65
& 21.74 & 0.582 & 0.3668 & 0.444 & 3.62 \\
DPS
& 29.35 & 0.828 & 0.2086 & 0.964 & 2.10
& \underline{25.23} & 0.696 & 0.2828 & 0.764 & 2.60
& 21.81 & 0.577 & \underline{0.3497} & 0.477 & 3.68 \\
DiffBIR
& 28.04 & 0.814 & 0.2153 & 0.968 & \underline{2.05}
& 23.85 & 0.675 & 0.3006 & 0.812 & \underline{2.46}
& 21.54 & 0.577 & 0.3650 & \underline{0.525} & \underline{3.44} \\
OSDFace
& 23.13 & 0.710 & 0.2470 & 0.951 & 2.24
& 20.27 & 0.590 & 0.3061 & \underline{0.825} & 2.90
& 17.80 & 0.470 & 0.3921 & 0.495 & 4.67 \\
FiDeSR
& 27.41 & 0.822 & \underline{0.1976} & 0.895 & 2.10
& 24.76 & 0.712 & \underline{0.2563} & 0.725 & 2.54
& \underline{21.98} & \underline{0.602} & 0.3619 & 0.463 & 4.68 \\
SubDAPS++
& \textbf{30.71} & \underline{0.852} & 0.2373 & \underline{0.971} & 2.07
& \textbf{26.90} & \textbf{0.750} & 0.3139 & 0.811 & 2.55
& \textbf{23.32} & \textbf{0.643} & 0.3720 & 0.513 & 3.51 \\
Ours
& \underline{30.00} & \textbf{0.863} & \textbf{0.1541} & \textbf{0.988} & \textbf{1.94}
& 25.12 & \underline{0.718} & \textbf{0.2559} & \textbf{0.859} & \textbf{2.39}
& 20.70 & 0.568 & \textbf{0.3482} & \textbf{0.553} & \textbf{3.34} \\
\bottomrule
\end{tabular}
}
\caption{Quantitative comparison of synthetic face super-resolution on the CelebA-HQ test set. Best and second-best results are marked in bold and underlined. ID~\cite{schroff2015facenet,cao2018vggface2} and LMD~\cite{bulat2017far} denote identity similarity and landmark distance.}
\label{tab:synthetic_results}
\end{table*}

\begin{figure*}[t]
\centering
\includegraphics[width=\textwidth]{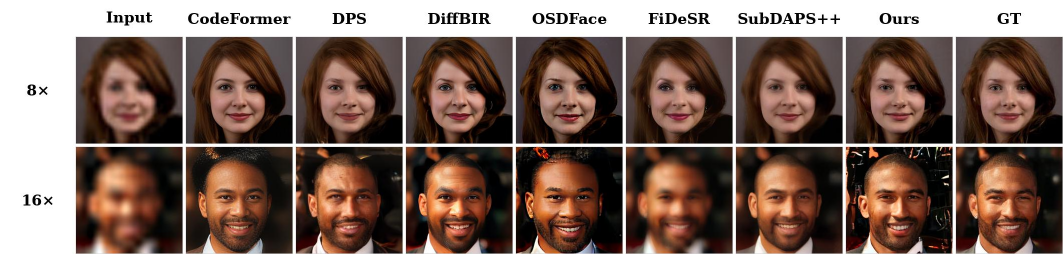}
\caption{Qualitative comparison on synthetic face super-resolution under \(8\times\) and \(16\times\) bicubic downsampling.}
\label{fig:qualitative_synthetic}
\end{figure*}

Starting from a noisy observation preserves the overall layout, but identity-related facial structure remains only indirectly constrained during reverse sampling. As shown in the lower-left panel of Fig.~\ref{fig:framework}, ASWG adds this constraint in a low-resolution anchor space, preserving coarse facial structure while reducing the influence of unreliable details. Let \(A(\cdot)\) denote anchor downsampling, whose output resolution matches the true LR size for synthetic SR and the selected anchor resolution for real-world restoration. At timestep \(t\), the clean estimate \(\hat{x}_{0|t}\) is projected to the anchor scale:
\begin{equation}
    u_t = A(\hat{x}_{0|t}).
\end{equation}
We set \(u_y=y^{lr}\) for synthetic SR and \(u_y=A(y)\) for real-world restoration. The pixel anchor loss is
\begin{equation}
    \mathcal{L}_{pixel} = \|u_t-u_y\|_2 .
\end{equation}
This term enforces point-wise consistency but does not explicitly describe structural differences. 

Inspired by wave-equation-based vision modeling~\cite{shu2026waveformer}, we apply the same fixed wave operator to \(u_t\) and \(u_y\), and use the response difference as a structural loss. The operator is not trained as a visual backbone and only provides a descriptor for sampling guidance.
To define the operator, let \(u\in\{u_t,u_y\}\) denote a generic anchor-space image. We compute a discrete Laplacian response and use it as the initial velocity of the wave dynamics:
\begin{equation}
    v_0 = \nabla_d^2 u = K_{\Delta} * u,\quad
    U=\mathcal{F}(u),\quad
    V=\mathcal{F}(v_0),
\end{equation}
where \(K_{\Delta}\) is a fixed channel-wise \(3\times3\) eight-neighbor kernel implementing the discrete Laplacian, \(v_0\) encodes local edge responses, and \(\mathcal{F}\) denotes the fast Fourier transform.
The damped anisotropic frequency is defined as
\begin{equation}
    \omega_d(k_x,k_y)=
    \sqrt{
    v_x^2 k_x^2 + v_y^2 k_y^2 - (\alpha/2)^2
    },
\end{equation}
where \(k_x\) and \(k_y\) denote spatial frequency coordinates, \(v_x\) and \(v_y\) are fixed propagation speeds along the two spatial directions, and \(\alpha\) is the damping coefficient. The term inside the square root is lower-bounded by a small positive constant for numerical stability. Different fixed speeds are used along the two directions to obtain a direction-dependent structural response.
The resulting wave response is written as
\begin{equation}
    \phi(u)=
    \mathcal{F}^{-1}
    \left[
    e^{-\alpha \tau/2}
    \left(
    U\cos(\omega_d \tau)
    +
    \frac{V+\alpha U/2}{\omega_d+\epsilon}
    \sin(\omega_d \tau)
    \right)
    \right],
\end{equation}
where \(\mathcal{F}^{-1}\) denotes the inverse fast Fourier transform, \(\tau\) is the propagation time, and \(\epsilon\) avoids division by zero.
We then define the wave-structural loss as
\begin{equation}
    \mathcal{L}_{wave}
    =
    \|\phi(u_t)-\phi(u_y)\|_2 .
\end{equation}
The total guidance objective used for posterior update is
\begin{equation}
    \mathcal{L}_{total}
    =
    \mathcal{L}_{pixel}
    +
    \lambda_t \mathcal{L}_{wave}.
\end{equation}
The gradient of \(\mathcal{L}_{total}\) is used in the posterior update. We pre-compute \(\phi(u_y)\) and stop the guidance at late timesteps to avoid over-constraining local details.

\begin{table*}[t]
\centering
{\small
\setlength{\tabcolsep}{1.6pt}
\begin{tabular}{l|ccccc|ccccc|ccccc}
\toprule
\multirow{2}{*}{Method}
& \multicolumn{5}{c|}{LFW-Test}
& \multicolumn{5}{c|}{WIDER-Test}
& \multicolumn{5}{c}{WebPhoto-Test} \\
& NIQE$\downarrow$ & PI$\downarrow$ & MUSIQ$\uparrow$ & C-IQA$\uparrow$ & TOPIQ$\uparrow$
& NIQE$\downarrow$ & PI$\downarrow$ & MUSIQ$\uparrow$ & C-IQA$\uparrow$ & TOPIQ$\uparrow$
& NIQE$\downarrow$ & PI$\downarrow$ & MUSIQ$\uparrow$ & C-IQA$\uparrow$ & TOPIQ$\uparrow$ \\
\midrule
CodeFormer
& 4.93 & 4.01 & 69.36 & 0.696 & 0.747
& 4.97 & 4.16 & 65.96 & 0.718 & 0.724
& 5.59 & 4.84 & 66.70 & 0.707 & 0.720 \\
DPS
& 6.84 & 6.41 & 53.50 & 0.578 & 0.615
& 11.34 & 9.61 & 19.74 & 0.300 & 0.210
& 9.29 & 8.36 & 31.60 & 0.386 & 0.382 \\
DiffBIR
& 5.36 & 4.44 & \underline{70.09} & \underline{0.724} & \underline{0.774}
& 5.73 & 4.93 & 67.59 & \underline{0.738} & 0.754
& 6.61 & 5.57 & 66.13 & 0.703 & 0.722 \\
OSDFace
& \underline{4.51} & \underline{3.42} & 68.55 & 0.719 & 0.770
& \underline{4.76} & \underline{3.67} & \underline{68.17} & 0.731 & \underline{0.763}
& \underline{4.90} & \underline{3.91} & \underline{67.00} & \underline{0.715} & \underline{0.739} \\
FiDeSR
& 5.00 & 4.04 & 69.45 & 0.716 & 0.769
& 6.20 & 5.43 & 63.58 & 0.718 & 0.721
& 6.47 & 5.56 & 66.38 & 0.694 & 0.738 \\
SubDAPS++
& 7.69 & 7.07 & 37.52 & 0.533 & 0.391
& 10.53 & 9.17 & 17.66 & 0.370 & 0.187
& 9.24 & 8.32 & 26.42 & 0.434 & 0.315 \\
Ours
& \textbf{4.36} & \textbf{2.80} & \textbf{72.77} & \textbf{0.762} & \textbf{0.808}
& \textbf{4.35} & \textbf{2.79} & \textbf{71.27} & \textbf{0.789} & \textbf{0.785}
& \textbf{4.71} & \textbf{3.44} & \textbf{71.30} & \textbf{0.756} & \textbf{0.773} \\
\bottomrule
\end{tabular}
}
\caption{Quantitative comparison on real-world face restoration. Best and second-best results are marked in bold and underlined. C-IQA denotes CLIPIQA~\cite{wang2023exploring}.}
\label{tab:real_results}
\end{table*}

\begin{figure*}[t]
\centering
\includegraphics[width=\textwidth]{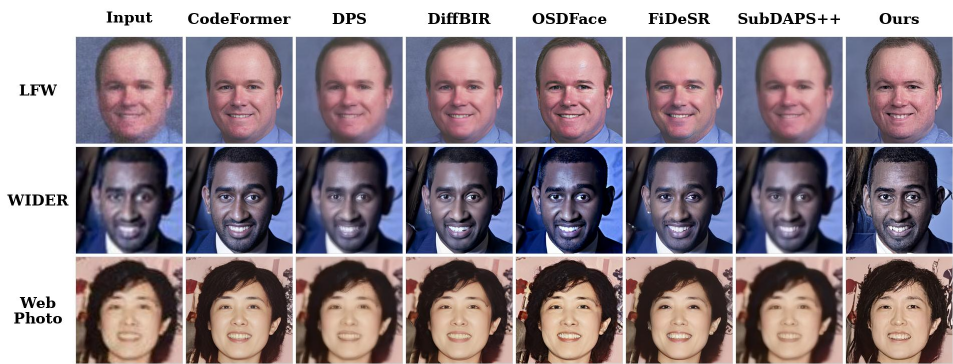}
\caption{Qualitative comparisons on real-world face restoration on LFW-Test, WIDER-Test, and WebPhoto-Test.}
\label{fig:qualitative_real}
\end{figure*}

\subsection{Multi-Scale Wavelet-Fourier Injection}

ASWG constrains facial structure in the anchor space. As shown in the upper panel of Fig.~\ref{fig:framework}, MWFI complements this guidance by using the observation phase to correct the predicted low-frequency subband while retaining the high-frequency subbands from the guided update.

The wavelet level \(J\) determines the resolution of the low-frequency subband used for correction. In synthetic SR, \(J=\log_2 s\) aligns this subband with the true LR size. For real-world restoration, \(J\) follows the selected sampling configuration, using a lower-resolution subband when fine-scale responses are weaker to reduce unreliable injection.

We apply a \(J\)-level Haar wavelet transform~\cite{mallat1989theory} to decompose an image \(z\) into a low-frequency subband and a set of high-frequency subbands:
\begin{equation}
    \left(LL_z,\{H_z^j\}_{j=1}^{J}\right)
    =
    \mathrm{DWT}_J(z),
\end{equation}
where \(LL_z\) captures the low-frequency structural content, and \(H_z^j\) contains the horizontal, vertical, and diagonal high-frequency subbands at level \(j\).
Before the reverse process, we take \(LL_y\) from the wavelet decomposition of the observation and compute its Fourier phase:
\begin{equation}
    \varphi_y = \angle \mathcal{F}(LL_y).
\end{equation}

At timestep \(t\), we decompose the current clean estimate \(\hat{x}_{0|t}\) and obtain its low-frequency Fourier representation:
\begin{equation}
    \mathcal{F}(LL_{\hat{x}_{0|t}})=A_t\exp(\mathrm{i}\varphi_t),
\end{equation}
where \(A_t\) and \(\varphi_t\) denote the predicted amplitude and phase. Since Fourier phase carries important spatial structure information~\cite{oppenheim1981importance}, we retain \(A_t\) and replace \(\varphi_t\) with the observation phase \(\varphi_y\):
\begin{equation}
    LL^{corr}
    =
    \mathcal{F}^{-1}
    \left(
    A_t \exp(\mathrm{i}\varphi_y)
    \right).
\end{equation}
The corrected low-frequency wavelet subband is then noised to the current timestep using the forward diffusion process:
\begin{equation}
    LL_t^{corr}
    =
    \sqrt{\bar{\alpha}_t}LL^{corr}
    +
    \sqrt{1-\bar{\alpha}_t}\varepsilon_t,
    \quad
    \varepsilon_t\sim\mathcal{N}(0,\mathbf{I}).
\end{equation}

Finally, we extract the high-frequency subbands from \(\tilde{x}_{t-1}\) after the posterior update and recombine them with \(LL_t^{corr}\):
\begin{equation}
    x_{t-1}
    =
    \mathrm{IDWT}_J
    \left(
    LL_t^{corr},
    \{H_{\tilde{x}_{t-1}}^j\}_{j=1}^{J}
    \right).
\end{equation}
This design injects observation-consistent low-frequency structure without overwriting the high-frequency subbands from the guided update. In practice, the injection is applied only in a middle timestep window, avoiding the very noisy early stage and leaving the final steps to refine details.

\subsection{Subband High-Frequency Enhancement}

For synthetic SR, the known downsampling model provides an explicit LR anchor, so we rely on the guided diffusion update without an additional refinement branch. Under unknown real-world degradations, high-frequency facial details may remain under-recovered in the predicted clean image. As shown in the lower-right panel of Fig.~\ref{fig:framework}, SHE refines only its high-frequency wavelet subbands rather than sharpening the final image in pixel space. SHE leaves the low-frequency subband unchanged because MWFI corrects it separately.

At timestep \(t\), we decompose the current clean estimate \(\hat{x}_{0|t}\) by the same \(J\)-level wavelet transform:
\begin{equation}
    (LL_{\hat{x}_{0|t}}, H_{\hat{x}_{0|t}})=\mathrm{DWT}_J(\hat{x}_{0|t}),
\end{equation}
where \(H_{\hat{x}_{0|t}}=\{H_{\hat{x}_{0|t}}^j\}_{j=1}^{J}\) denotes the high-frequency subbands at different levels. For each level, we bound the high-frequency response to avoid amplifying unstable artifacts:
\begin{equation}
    \bar{H}_{\hat{x}_{0|t}}^j
    =
    \mathrm{clip}
    \left(
    H_{\hat{x}_{0|t}}^j,
    -k,
    k
    \right),
\end{equation}
where \(k>0\) is a fixed clipping threshold shared across wavelet levels. We then apply a controlled gain:
\begin{equation}
    \widetilde{H}_{\hat{x}_{0|t}}^j
    =
    H_{\hat{x}_{0|t}}^j
    +
    \gamma_t M^j \odot \bar{H}_{\hat{x}_{0|t}}^j ,
\end{equation}
where \(\gamma_t\) is a boosting factor that is linearly reduced during the active refinement window, \(M^j\) is the spatial mask at level \(j\), and \(\odot\) denotes the Hadamard product. We use an elliptical spatial mask that assigns larger weights near the central face region and rapidly attenuates toward the boundary, reducing high-frequency amplification in background areas.

The result is reconstructed by inverse wavelet transform:
\begin{equation}
    \hat{x}_{0|t}^{*}
    =
    \mathrm{IDWT}_J
    \left(
    LL_{\hat{x}_{0|t}},
    \{\widetilde{H}_{\hat{x}_{0|t}}^j\}_{j=1}^{J}
    \right).
\end{equation}
\(\hat{x}_{0|t}^{*}\) is used to compute the posterior mean in the guided reverse update. This refinement strengthens local high-frequency details while leaving the low-frequency subband unchanged and introduces no additional generative prior.
\section{Experiments}

\subsection{Experimental Settings}

\begin{table}[t]
\centering
\small
\begin{tabular}{lcc}
\toprule
Method / Variant & ID$\uparrow$ & LMD$\downarrow$ \\
\midrule
DPS~\cite{chung2023diffusion} & 0.4725 & 3.3065 \\
RED-Diff~\cite{mardani2024variational} & \underline{0.5798} & 3.0898 \\
DAPS~\cite{zhang2025improving} & 0.4585 & 3.9058 \\
SubDAPS++~\cite{zheng2026image} & 0.5207 & 3.4783 \\
\midrule
Pixel anchor only & 0.5531 & 3.2145 \\
Anchor-space wave-structural guidance & 0.5735 & \underline{3.0777} \\
+ Multi-scale wavelet-Fourier injection & \textbf{0.5852} & \textbf{3.0472} \\
\bottomrule
\end{tabular}
\caption{Component ablation and comparison with observation-guided diffusion methods on \(16\times\) FSR.}
\label{tab:ablation_synthetic}
\end{table}

\begin{table}[t]
\centering
\small
\begin{tabular}{lccc}
\toprule
Initial velocity & Propagation & ID$\uparrow$ & LMD$\downarrow$ \\
\midrule
\(v_0=0\)
& Anisotropic
& \textbf{0.5789}
& 3.1268 \\
\(v_0=\nabla_d^2u\)
& Isotropic
& 0.5706
& \textbf{3.0650} \\
\(v_0=\nabla_d^2u\)
& Anisotropic
& \underline{0.5735}
& \underline{3.0777} \\
\bottomrule
\end{tabular}
\caption{Ablation of \(\phi(\cdot)\) on \(16\times\) face super-resolution. The pixel anchor is retained and MWFI is disabled. }
\label{tab:ablation_phi}
\end{table}

\begin{table}[t]
\centering
\small
\begin{tabular}{@{}lccc@{}}
\toprule
State
& PSNR (dB) $\uparrow$
& LPIPS $\downarrow$
& DISTS $\downarrow$ \\
\midrule
Before correction
& 25.03
& 0.1555
& 0.0733 \\
After correction
& \textbf{25.84}
& \textbf{0.1480}
& \textbf{0.0727} \\
\bottomrule
\end{tabular}
\caption{Low-frequency phase-correction analysis of MWFI under \(16\times\) FSR. The LL-only reconstructions before and after correction are compared against ground truth.}
\label{tab:mwfi_analysis}
\end{table}

\begin{figure}[t]
\centering
\includegraphics[width=\columnwidth]
{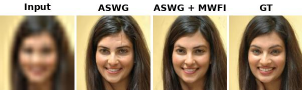}
\caption{Qualitative effect of MWFI under \(16\times\) SR. }
\label{fig:mwfi_visual}
\end{figure}

\begin{table}[t]
\centering
\small
\setlength{\tabcolsep}{3.5pt}
\begin{tabular}{lccccc}
\toprule
Variant & NIQE$\downarrow$ & PI$\downarrow$ & MUSIQ$\uparrow$ & C-IQA$\uparrow$ & TOPIQ$\uparrow$ \\
\midrule
w/o SHE & 5.899 & 5.248 & 63.76 & 0.657 & 0.717 \\
Full & \textbf{4.535} & \textbf{3.028} & \textbf{71.72} & \textbf{0.769} & \textbf{0.786} \\
\bottomrule
\end{tabular}
\caption{Ablation of our SHE on real-world test sets.}
\label{tab:ablation_she}
\end{table}

\begin{figure}[t]
\centering
\includegraphics[width=\columnwidth]
{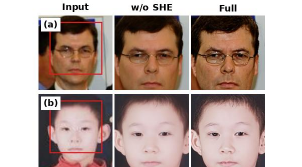}
\caption{Qualitative effect of SHE on real-world restoration.}
\label{fig:she_visual}
\end{figure}

\begin{figure}[t]
\centering
\includegraphics[width=\columnwidth]{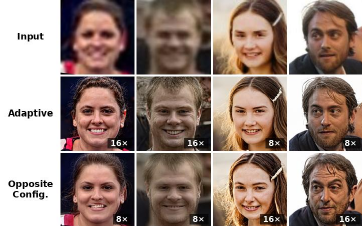}
\caption{The configuration selected by our adaptive strategy consistently produces superior visual quality compared with the alternatives. The label on each image indicates the selected predefined scale configuration.}
\label{fig:router_ablation}
\end{figure}

Our method uses a frozen unconditional diffusion model~\cite{chung2023diffusion} pretrained on FFHQ~\cite{karras2019style}. We evaluate synthetic face super-resolution on a fixed 1000-image subset of CelebA-HQ~\cite{karras2018progressive} at \(4\times\), \(8\times\), and \(16\times\). Each high-quality image is resized to \(256\times256\), downsampled by bicubic interpolation with anti-aliasing, and upsampled back to \(256\times256\) as low-quality images. Real-world evaluation uses LFW-Test, WIDER-Test, and WebPhoto-Test provided by Codeformer~\cite{zhou2022towards}.

We compare with CodeFormer~\cite{zhou2022towards}, DPS~\cite{chung2023diffusion}, DiffBIR~\cite{lin2024diffbir}, OSDFace~\cite{wang2025osdface}, FiDeSR~\cite{kim2026fidesr} and SubDAPS++~\cite{zheng2026image}. All methods use the same bicubic LR measurement for synthetic SR and the same aligned \(256\times256\) input for real-world restoration. Pipelines operating at \(512\times512\) resize only the input internally, and their outputs are resized to \(256\times256\) before evaluation. Synthetic metrics are PSNR, SSIM~\cite{wang2004image}, LPIPS~\cite{zhang2018unreasonable}, ID cosine similarity between Inception-ResNet embeddings pretrained on VGGFace2~\cite{schroff2015facenet,cao2018vggface2}, and FAN landmark RMS distance~\cite{bulat2017far}. Real-world metrics are NIQE~\cite{mittal2013making}, PI~\cite{blau2018pirm}, MUSIQ~\cite{ke2021musiq}, CLIPIQA~\cite{wang2023exploring}, and TOPIQ~\cite{chen2024topiq}. 

We use a 1000-step diffusion process with a linear noise schedule. For \(4\times\), \(8\times\), and \(16\times\) synthetic SR, the starting timesteps are 500, 700, and 800. The anchor resolutions are \(64\times64\), \(32\times32\), and \(16\times16\), matching the true LR sizes, and the corresponding Haar wavelet levels are \(J=2,3,4\). The guidance strengths are \((\zeta,\lambda)=(2.5,2.0),(3.0,2.5),(3.5,3.0)\), which are kept fixed during synthetic sampling. For real-world inputs, the frequency-ratio router selects the synthetic \(8\times\) or \(16\times\) parameter setting using \(\tau_r=0.065\) with linear annealing of the guidance strengths at late timesteps. Experiments run on an NVIDIA RTX 5060 Ti (16GB) with PyTorch under Ubuntu 24.04.

\subsection{Comparison with State-of-the-Art Methods}

Table~\ref{tab:synthetic_results} compares synthetic FSR under \(4\times\), \(8\times\), and \(16\times\). At \(4\times\), our method ranks first on SSIM, LPIPS, ID, and LMD, and second on PSNR. At \(8\times\) and \(16\times\), it obtains the best LPIPS, ID, and LMD. The advantages in identity similarity and landmark consistency remain evident under severe \(16\times\) downsampling. The visual comparisons in Figure ~\ref{fig:qualitative_synthetic} show that our method can reconstruct faces with features that are closer to ground truth. Table~\ref{tab:real_results} and Fig.~\ref{fig:qualitative_real} provide the quantitative and qualitative results on real-world scenes. Our method obtains the best results on all reported metrics across LFW-Test, WIDER-Test, and WebPhoto-Test, and achieves strong perceptual quality under real-world degradations.

\subsection{Ablation Study}

\paragraph{Component contributions.}
The component ablation focuses on the severe \(16\times\) setting, where identity-related structural drift is more evident. The last three rows of Table~\ref{tab:ablation_synthetic} show that ASWG improves identity similarity and landmark consistency over pixel-anchor-only guidance. Adding MWFI further improves both aspects. These results show that the two components progressively reduce identity-related structural drift under severe downsampling.

\paragraph{Definition of \(\phi(\cdot)\).}
We examine the Laplacian initialization and direction-dependent propagation used in \(\phi(\cdot)\). As shown in Table~\ref{tab:ablation_phi}, removing the Laplacian initialization improves identity similarity but weakens landmark consistency, whereas isotropic propagation slightly improves landmark consistency at the cost of identity similarity. The adopted Laplacian-initialized anisotropic form provides a better balance between identity similarity and landmark consistency.

\paragraph{Comparison with observation-guided methods.}
Table~\ref{tab:ablation_synthetic} also compares ASWG with representative observation-guided diffusion methods. ASWG achieves the best landmark consistency among the compared methods while maintaining competitive identity similarity, showing that its structural guidance more reliably preserves facial geometry.

\paragraph{Low-frequency correction.}
At each timestep in the active MWFI window, we compare the LL-only reconstructions before and after phase correction with their ground-truth counterparts using PSNR, LPIPS, and DISTS~\cite{ding2022image}, with all high-frequency subbands set to zero. As shown in Table~\ref{tab:mwfi_analysis}, phase correction improves both the pixel fidelity and perceptual similarity of the low-frequency reconstruction. Fig.~\ref{fig:mwfi_visual} further shows that MWFI better aligns the overall facial shape and the relative positions of facial components with the ground truth in the final output.

\paragraph{High-frequency enhancement.}
Because SHE targets high-frequency facial details that may remain under-recovered in real-world inputs, we evaluate it by disabling the high-frequency subband refinement. As shown in Table~\ref{tab:ablation_she}, SHE improves the overall no-reference perceptual quality of real-world restoration. Fig.~\ref{fig:she_visual} further shows clearer eyeglass frames, hair boundaries, and local facial contours, supporting its ability to recover high-frequency facial details.

\paragraph{Adaptive configuration.}
A single sampling configuration is not suitable for all real-world inputs. As shown in Fig.~\ref{fig:router_ablation}, the \(16\times\) preset produces clearer facial components when fine-scale cues are weak. When more usable cues remain, the \(8\times\) preset better retains the input-supported facial appearance, whereas the \(16\times\) preset produces more pronounced changes in local facial characteristics. The frequency-ratio router therefore selects between the two predefined configurations according to \(r_y\).

\subsection{Limitations and Future Work}
For our method, severe color casts, over-exposure, and low facial contrast also remain challenging. The two-configuration router cannot fully cover the continuous range of real degradations. Future work will explore robust identity-related constraints and degradation-aware parameter selection.

\section{Conclusion}
We proposed WaveFreqAnchor, a training-free framework that improves diffusion-based face restoration through wave-structural anchoring and frequency-guided correction. By jointly introducing ASWG, MWFI, and SHE, our method effectively mitigates structural drift while preserving facial identity and recovering realistic high-frequency details under severe real-world degradations. Extensive experiments demonstrate consistent improvements over state-of-the-art methods, validating the effectiveness of combining wavelet-domain structural guidance with frequency-aware diffusion for high-fidelity face restoration.

\bibliography{ref}

@inproceedings{ho2020denoising,
  author    = {Ho, Jonathan and Jain, Ajay and Abbeel, Pieter},
  title     = {Denoising Diffusion Probabilistic Models},
  booktitle = {Advances in Neural Information Processing Systems},
  volume    = {33},
  pages     = {6840--6851},
  year      = {2020},
}

@inproceedings{kawar2022denoising,
  author    = {Kawar, Bahjat and Elad, Michael and Ermon, Stefano and Song, Jiaming},
  title     = {Denoising Diffusion Restoration Models},
  booktitle = {Advances in Neural Information Processing Systems},
  volume    = {35},
  pages     = {23593--23606},
  year      = {2022},
}

@inproceedings{chung2023diffusion,
  author    = {Chung, Hyungjin and Kim, Jeongsol and McCann, Michael T. and Klasky, Marc L. and Ye, Jong Chul},
  title     = {Diffusion Posterior Sampling for General Noisy Inverse Problems},
  booktitle = {Proceedings of the International Conference on Learning Representations},
  year      = {2023}
}

@inproceedings{meng2022sdedit,
  author    = {Meng, Chenlin and He, Yutong and Song, Yang and Song, Jiaming and Wu, Jiajun and Zhu, Jun-Yan and Ermon, Stefano},
  title     = {{SDEdit}: Guided Image Synthesis and Editing with Stochastic Differential Equations},
  booktitle = {Proceedings of the International Conference on Learning Representations},
  year      = {2022}
}

@inproceedings{wang2021towards,
  author    = {Wang, Xintao and Li, Yu and Zhang, Honglun and Shan, Ying},
  title     = {Towards Real-World Blind Face Restoration with Generative Facial Prior},
  booktitle = {Proceedings of the IEEE/CVF Conference on Computer Vision and Pattern Recognition},
  pages     = {9168--9178},
  year      = {2021}
}

@inproceedings{gu2022vqfr,
  author    = {Gu, Yuchao and Wang, Xintao and Xie, Liangbin and Dong, Chao and Li, Gen and Shan, Ying and Cheng, Ming-Ming},
  title     = {{VQFR}: Blind Face Restoration with Vector-Quantized Dictionary and Parallel Decoder},
  booktitle = {Proceedings of the European Conference on Computer Vision},
  volume    = {13678},
  pages     = {126--143},
  year      = {2022}
}

@inproceedings{zhou2022towards,
  author    = {Zhou, Shangchen and Chan, Kelvin C. K. and Li, Chongyi and Loy, Chen Change},
  title     = {Towards Robust Blind Face Restoration with Codebook Lookup Transformer},
  booktitle = {Advances in Neural Information Processing Systems},
  volume    = {35},
  pages     = {30599--30611},
  year      = {2022}
}

@inproceedings{lin2024diffbir,
  author    = {Lin, Xinqi and He, Jingwen and Chen, Ziyan and
               Lyu, Zhaoyang and Dai, Bo and Yu, Fanghua and
               Qiao, Yu and Ouyang, Wanli and Dong, Chao},
  title     = {{DiffBIR}: Toward Blind Image Restoration with
               Generative Diffusion Prior},
  booktitle = {Proceedings of the European Conference on Computer Vision},
  volume    = {15117},
  pages     = {430--448},
  year      = {2024}
}

@inproceedings{shu2026waveformer,
  author    = {Shu, Zishan and Wu, Juntong and Yan, Wei and
               Liu, Xudong and Zhang, Hongyu and Liu, Chang and
               Mao, Youdong and Chen, Jie},
  title     = {{WaveFormer}: Frequency-Time Decoupled Vision Modeling
               with Wave Equation},
  booktitle = {Proceedings of the AAAI Conference on Artificial Intelligence},
  volume    = {40},
  pages     = {25428--25436},
  year      = {2026}
}

@article{mallat1989theory,
  author    = {Mallat, S.G.},
  title     = {A theory for multiresolution signal decomposition: the wavelet representation},
  journal   = {IEEE Transactions on Pattern Analysis and Machine Intelligence},
  volume    = {11},
  number    = {7},
  pages     = {674--693},
  year      = {1989},
  doi       = {10.1109/34.192463}
}

@article{oppenheim1981importance,
  author    = {Oppenheim, Alan V. and Lim, Jae S.},
  title     = {The importance of phase in signals},
  journal   = {Proceedings of the IEEE},
  volume    = {69},
  number    = {5},
  pages     = {529--541},
  year      = {1981},
  doi       = {10.1109/PROC.1981.12022}
}

@inproceedings{karras2018progressive,
  author    = {Karras, Tero and Aila, Timo and Laine, Samuli and Lehtinen, Jaakko},
  title     = {Progressive Growing of {GANs} for Improved Quality, Stability, and Variation},
  booktitle = {Proceedings of the International Conference on Learning Representations},
  year      = {2018}
}

@article{li2025survey,
  title={Survey on deep face restoration: From non-blind to blind and beyond},
  author={Li, Wenjie and Wang, Mei and Zhang, Kai and Li, Juncheng and Li, Xiaoming and Zhang, Yuhang and Gao, Guangwei and Ma, Zhanyu},
  journal={ACM Computing Surveys},
  year={2025},
  publisher={ACM New York, NY}
}

@inproceedings{wang2025osdface,
  author    = {Wang, Jingkai and Gong, Jue and Zhang, Lin and Chen, Zheng and Liu, Xing and Gu, Hong and Liu, Yutong and Zhang, Yulun and Yang, Xiaokang},
  title     = {{OSDFace}: One-Step Diffusion Model for Face Restoration},
  booktitle = {Proceedings of the IEEE/CVF Conference on Computer Vision and Pattern Recognition},
  pages     = {12626--12636},
  year      = {2025},
  doi       = {10.1109/CVPR52734.2025.01178}
}

@article{wang2004image,
  author    = {Wang, Zhou and Bovik, A.C. and Sheikh, H.R. and Simoncelli, E.P.},
  title     = {Image Quality Assessment: From Error Visibility to Structural Similarity},
  journal   = {IEEE Transactions on Image Processing},
  volume    = {13},
  number    = {4},
  pages     = {600--612},
  year      = {2004},
  doi       = {10.1109/TIP.2003.819861}
}

@inproceedings{zhang2018unreasonable,
  author    = {Zhang, Richard and Isola, Phillip and Efros, Alexei A. and Shechtman, Eli and Wang, Oliver},
  title     = {The Unreasonable Effectiveness of Deep Features as a Perceptual Metric},
  booktitle = {Proceedings of the IEEE Conference on Computer Vision and Pattern Recognition},
  pages     = {586--595},
  year      = {2018},
  doi       = {10.1109/CVPR.2018.00068}
}

@inproceedings{schroff2015facenet,
  author    = {Schroff, Florian and Kalenichenko, Dmitry and Philbin, James},
  title     = {{FaceNet}: A unified embedding for face recognition and clustering},
  booktitle = {Proceedings of the IEEE Conference on Computer Vision and Pattern Recognition},
  pages     = {815--823},
  year      = {2015},
  doi       = {10.1109/CVPR.2015.7298682}
}

@inproceedings{cao2018vggface2,
  author    = {Cao, Qiong and Shen, Li and Xie, Weidi and Parkhi, Omkar M. and Zisserman, Andrew},
  title     = {{VGGFace2}: A Dataset for Recognising Faces across Pose and Age},
  booktitle = {Proceedings of the IEEE International Conference on Automatic Face \& Gesture Recognition (FG 2018)},
  pages     = {67--74},
  year      = {2018},
  doi       = {10.1109/FG.2018.00020}
}

@inproceedings{bulat2017far,
  author    = {Bulat, Adrian and Tzimiropoulos, Georgios},
  title     = {How Far Are We from Solving the {2D} \& {3D} Face Alignment Problem? (and a Dataset of 230,000 {3D} Facial Landmarks)},
  booktitle = {Proceedings of the IEEE International Conference on Computer Vision},
  pages     = {1021--1030},
  year      = {2017},
  doi       = {10.1109/ICCV.2017.116}
}

@article{mittal2013making,
  author    = {Mittal, Anish and Soundararajan, Rajiv and Bovik, Alan C.},
  title     = {Making a ``Completely Blind'' Image Quality Analyzer},
  journal   = {IEEE Signal Processing Letters},
  volume    = {20},
  number    = {3},
  pages     = {209--212},
  year      = {2013},
  doi       = {10.1109/LSP.2012.2227726}
}

@inproceedings{li2024efficient,
  title={Efficient face super-resolution via wavelet-based feature enhancement network},
  author={Li, Wenjie and Guo, Heng and Liu, Xuannan and Liang, Kongming and Hu, Jiani and Ma, Zhanyu and Guo, Jun},
  booktitle={Proceedings of the ACM International Conference on Multimedia},
  pages={4515--4523},
  year={2024}
}

@inproceedings{blau2018pirm,
  author    = {Blau, Yochai and Mechrez, Roey and Timofte, Radu and Michaeli, Tomer and Zelnik-Manor, Lihi},
  title     = {The 2018 {PIRM} Challenge on Perceptual Image Super-resolution},
  booktitle = {Proceedings of the European Conference on Computer Vision Workshops},
  series    = {Lecture Notes in Computer Science},
  volume    = {11133},
  pages     = {334--355},
  year      = {2019},
  publisher = {Springer},
  doi       = {10.1007/978-3-030-11021-5_21}
}

@inproceedings{ke2021musiq,
  author    = {Ke, Junjie and Wang, Qifei and Wang, Yilin and Milanfar, Peyman and Yang, Feng},
  title     = {{MUSIQ}: Multi-scale Image Quality Transformer},
  booktitle = {Proceedings of the IEEE/CVF International Conference on Computer Vision},
  pages     = {5148--5157},
  year      = {2021},
  doi       = {10.1109/ICCV48922.2021.00510}
}

@inproceedings{wang2023exploring,
  author    = {Wang, Jianyi and Chan, Kelvin C. K. and Loy, Chen Change},
  title     = {Exploring {CLIP} for Assessing the Look and Feel of Images},
  booktitle = {Proceedings of the AAAI Conference on Artificial Intelligence},
  volume    = {37},
  pages     = {2555--2563},
  year      = {2023},
  doi       = {10.1609/aaai.v37i2.25353}
}

@article{chen2024topiq,
  author    = {Chen, Chaofeng and Mo, Jiadi and Hou, Jingwen and Wu, Haoning and Liao, Liang and Sun, Wenxiu and Yan, Qiong and Lin, Weisi},
  title     = {{TOPIQ}: A Top-Down Approach From Semantics to Distortions for Image Quality Assessment},
  journal   = {IEEE Transactions on Image Processing},
  volume    = {33},
  pages     = {2404--2418},
  year      = {2024},
  doi       = {10.1109/TIP.2024.3378466}
}

@inproceedings{kim2026fidesr,
  author    = {Kim, Aro and Jang, Myeongjin and Moon, Chaewon and
               Shin, Youngjin and Jeong, Jinwoo and Park, Sang-hyo},
  title     = {{FiDeSR}: High-Fidelity and Detail-Preserving One-Step
               Diffusion Super-Resolution},
  booktitle = {Proceedings of the IEEE/CVF Conference on Computer Vision and Pattern Recognition},
  pages     = {38270--38280},
  year      = {2026}
}

@inproceedings{zheng2026image,
  author    = {Zheng, Yang and Li, Wen and Liu, Zhaoqiang},
  title     = {Image Restoration via Diffusion Models with Dynamic Resolution},
  booktitle = {Proceedings of the International Conference on Machine Learning},
  year      = {2026}
}

@inproceedings{menon2020pulse,
  author    = {Menon, Sachit and Damian, Alexandru and Hu, Shijia and
               Ravi, Nikhil and Rudin, Cynthia},
  title     = {{PULSE}: Self-Supervised Photo Upsampling via Latent Space
               Exploration of Generative Models},
  booktitle = {Proceedings of the IEEE/CVF Conference on Computer Vision and Pattern Recognition},
  pages     = {2434--2442},
  year      = {2020}
}

@inproceedings{wang2023zeroshot,
  author    = {Wang, Yinhuai and Yu, Jiwen and Zhang, Jian},
  title     = {Zero-Shot Image Restoration Using Denoising Diffusion
               Null-Space Model},
  booktitle = {Proceedings of the International Conference on Learning Representations},
  year      = {2023}
}

@inproceedings{mardani2024variational,
  author    = {Mardani, Morteza and Song, Jiaming and
               Kautz, Jan and Vahdat, Arash},
  title     = {A Variational Perspective on Solving Inverse Problems
               with Diffusion Models},
  booktitle = {Proceedings of the International Conference on Learning Representations},
  year      = {2024}
}

@inproceedings{zhang2025improving,
  author    = {Zhang, Bingliang and Chu, Wenda and Berner, Julius and
               Meng, Chenlin and Anandkumar, Anima and Song, Yang},
  title     = {Improving Diffusion Inverse Problem Solving with
               Decoupled Noise Annealing},
  booktitle = {Proceedings of the IEEE/CVF Conference on Computer Vision and Pattern Recognition},
  pages     = {20895--20905},
  year      = {2025}
}

@inproceedings{jiang2021focal,
  author    = {Jiang, Liming and Dai, Bo and Wu, Wayne and Loy, Chen Change},
  title     = {Focal Frequency Loss for Image Reconstruction and
               Synthesis},
  booktitle = {Proceedings of the IEEE/CVF International Conference on Computer Vision},
  pages     = {13919--13929},
  year      = {2021}
}

@inproceedings{wang2022restoreformer,
  author    = {Wang, Zhouxia and Zhang, Jiawei and Chen, Runjian and
               Wang, Wenping and Luo, Ping},
  title     = {{RestoreFormer}: High-Quality Blind Face Restoration from
               Undegraded Key-Value Pairs},
  booktitle = {Proceedings of the IEEE/CVF Conference on Computer Vision and Pattern Recognition},
  pages     = {17512--17521},
  year      = {2022}
}

@inproceedings{xie2024pltrans,
  author    = {Xie, Lianxin and Zheng, Bingbing and Xue, Wen and Jiang, Le
               and Liu, Cheng and Wu, Si and Wong, Hau-San},
  title     = {Learning Degradation-Unaware Representation with Prior-Based
               Latent Transformations for Blind Face Restoration},
  booktitle = {Proceedings of the IEEE/CVF Conference on Computer Vision and Pattern Recognition},
  pages     = {9120--9129},
  year      = {2024},
  doi       = {10.1109/CVPR52733.2024.00871}
}

@inproceedings{wang2023dr2,
  author    = {Wang, Zhixin and Zhang, Ziying and Zhang, Xiaoyun and
               Zheng, Huangjie and Zhou, Mingyuan and Zhang, Ya and
               Wang, Yanfeng},
  title     = {{DR2}: Diffusion-Based Robust Degradation Remover for Blind
               Face Restoration},
  booktitle = {Proceedings of the IEEE/CVF Conference on Computer Vision and Pattern Recognition},
  pages     = {1704--1713},
  year      = {2023}
}

@article{yue2024difface,
  author  = {Yue, Zongsheng and Loy, Chen Change},
  title   = {{DifFace}: Blind Face Restoration with Diffused Error
             Contraction},
  journal = {IEEE Transactions on Pattern Analysis and Machine Intelligence},
  volume  = {46},
  number  = {12},
  pages   = {9991--10004},
  year    = {2024},
  doi     = {10.1109/TPAMI.2024.3432651}
}

@inproceedings{do2025dynfacerestore,
  author    = {Do, Huu-Phu and Chen, Yu-Wei and Liao, Yi-Cheng and Hsiao, Chi-Wei and Wang, Han-Yang and Chiu, Wei-Chen and Huang, Ching-Chun},
  title     = {{DynFaceRestore}: Balancing Fidelity and Quality in Diffusion-Guided Blind Face Restoration with Dynamic Blur-Level Mapping and Guidance},
  booktitle = {Proceedings of the IEEE/CVF International Conference on Computer Vision},
  pages     = {10432--10441},
  year      = {2025}
}

@inproceedings{yang2023pgdiff,
  author    = {Yang, Peiqing and Zhou, Shangchen and Tao, Qingyi and
               Loy, Chen Change},
  title     = {{PGDiff}: Guiding Diffusion Models for Versatile Face
               Restoration via Partial Guidance},
  booktitle = {Advances in Neural Information Processing Systems},
  volume    = {36},
  year      = {2023}
}

@inproceedings{li2025self,
  author    = {Li, Wenjie and Wang, Xiangyi and Guo, Heng and
               Gao, Guangwei and Ma, Zhanyu},
  title     = {Self-Supervised Selective-Guided Diffusion Model for
               Old-Photo Face Restoration},
  booktitle = {Advances in Neural Information Processing Systems},
  year      = {2025}
}

@inproceedings{chung2024prompt,
  author    = {Chung, Hyungjin and Ye, Jong Chul and Milanfar, Peyman and
               Delbracio, Mauricio},
  title     = {Prompt-Tuning Latent Diffusion Models for Inverse Problems},
  booktitle = {Proceedings of the 41st International Conference on Machine
               Learning},
  series    = {Proceedings of Machine Learning Research},
  volume    = {235},
  pages     = {8941--8967},
  year      = {2024},
  publisher = {PMLR}
}

@inproceedings{yang2021gpen,
  author    = {Yang, Tao and Ren, Peiran and Xie, Xuansong and Zhang, Lei},
  title     = {{GAN} Prior Embedded Network for Blind Face Restoration in
               the Wild},
  booktitle = {Proceedings of the IEEE/CVF Conference on Computer Vision and Pattern Recognition},
  pages     = {672--681},
  year      = {2021}
}

@inproceedings{fuoli2021fourier,
  author    = {Fuoli, Dario and Van Gool, Luc and Timofte, Radu},
  title     = {Fourier Space Losses for Efficient Perceptual Image
               Super-Resolution},
  booktitle = {Proceedings of the IEEE/CVF International Conference on Computer Vision},
  pages     = {2360--2369},
  year      = {2021}
}

@inproceedings{miao2024waveface,
  author    = {Miao, Yunqi and Deng, Jiankang and Han, Jungong},
  title     = {{WaveFace}: Authentic Face Restoration with Efficient
               Frequency Recovery},
  booktitle = {Proceedings of the IEEE/CVF Conference on Computer Vision and Pattern Recognition},
  pages     = {6583--6592},
  year      = {2024}
}

@inproceedings{yang2026hdwsr,
  author    = {Yang, Chao and Zhang, Boqian and Xu, Jinghao and Jiang, Guang},
  title     = {{HDW-SR}: High-Frequency Guided Diffusion Model Based on
               Wavelet Decomposition for Image Super-Resolution},
  booktitle = {Proceedings of the IEEE/CVF Conference on Computer Vision and Pattern Recognition},
  pages     = {23462--23472},
  year      = {2026}
}

@inproceedings{choi2026framer,
  author    = {Choi, Seungho and Sung, Jeahun and Oh, Jihyong},
  title     = {{FRAMER}: Frequency-Aligned Self-Distillation with Adaptive
               Modulation Leveraging Diffusion Priors for Real-World Image
               Super-Resolution},
  booktitle = {Proceedings of the IEEE/CVF Conference on Computer Vision and Pattern Recognition},
  pages     = {23451--23461},
  year      = {2026}
}

@inproceedings{li2026seeing,
  author    = {Li, Wenjie and Shi, Jinglei and Han, Jin and Guo, Heng and
               Ma, Zhanyu},
  title     = {Seeing Through the Rain: Resolving High-Frequency Conflicts
               in Deraining and Super-Resolution via Diffusion Guidance},
  booktitle = {Proceedings of the AAAI Conference on Artificial Intelligence},
  volume    = {40},
  pages     = {6468--6476},
  year      = {2026},
  doi       = {10.1609/aaai.v40i8.37575}
}

@inproceedings{ni2026pasdiff,
  title={PASDiff: Physics-Aware Semantic Guidance for Joint Real-World
  Low-Light Face Enhancement and Restoration},
  author={Ni, Yilin and Li, Wenjie and Wang, Zhengxue and Li, Juncheng
  and Gao, Guangwei and Yang, Jian},
  booktitle={Proceedings of the European Conference on Computer Vision},
  year={2026}
}

@inproceedings{karras2019style,
  author    = {Karras, Tero and Laine, Samuli and Aila, Timo},
  title     = {A Style-Based Generator Architecture for Generative
               Adversarial Networks},
  booktitle = {Proceedings of the IEEE/CVF Conference on Computer Vision and Pattern Recognition},
  pages     = {4396--4405},
  year      = {2019},
  doi       = {10.1109/CVPR.2019.00453}
}

@article{ding2022image,
  author  = {Ding, Keyan and Ma, Kede and Wang, Shiqi and
             Simoncelli, Eero P.},
  title   = {Image Quality Assessment: Unifying Structure and
             Texture Similarity},
  journal = {IEEE Transactions on Pattern Analysis and Machine Intelligence},
  volume  = {44},
  number  = {5},
  pages   = {2567--2581},
  year    = {2022},
  doi     = {10.1109/TPAMI.2020.3045810}
}


\end{document}